\documentclass[twocolumn]{autart} 
\usepackage{amsmath,amssymb,euscript,yfonts,psfrag,latexsym,graphicx}
\usepackage{bbm,color,amstext,wasysym,parskip}

\graphicspath{{./},{../figures/},{./figures/}}
\usepackage{caption}
\usepackage{subcaption}
\usepackage{float}
\usepackage{array}
\usepackage{url}
\usepackage{algorithm}
\usepackage{algorithmic}
\usepackage{xcolor}

\usepackage{hyperref}

\graphicspath{{./},{../figures/},{./figures/}}

\usepackage{soul}
\usepackage[normalem]{ulem}

\newcommand{\bn}{{\mathbf n}}
\newcommand{\by}{{\mathbf y}}

\newcommand{\bo}{{\mathbf o}}

\newcommand{\singh}[1]{\textcolor{black}{#1}}

\newcommand{\cF}{{\mathcal F}}

\newcommand{\cL}{{\mathcal L}}

\newcommand{\cO}{{\mathcal O}}

\newcommand{\cX}{{\mathcal X}}

\newcommand{\bx}{{\bf x}}

\def\minwrt[#1]{\underset{#1}{\text{minimize}}}
\def\maxwrt[#1]{\underset{#1}{\text{maximize }}}

\newenvironment{proof}[1][Proof]{\noindent\textbf{#1.} }{\ \rule{0.5em}{0.5em}}

\newcommand{\argmin}{\operatorname{argmin}}

\usepackage{algorithmic}

\floatname{algorithm}{Algorithm}

\newlength{\bibitemsep}
\newlength{\bibparskip}
\let\oldthebibliography\thebibliography
\renewcommand\thebibliography[1]{%
  \oldthebibliography{#1}%
  \setlength{\parskip}{\bibitemsep}%
  \setlength{\itemsep}{\bibparskip}%
}
\begin{document}

\begin{frontmatter}

\title{Learning Hidden Markov Models from Aggregate Observations}

\thanks[footnoteinfo]{This work was supported by the NSF under grant 1901599, 1942523 and 2008513.}

\author[rahul]{Rahul Singh}\ead{rasingh@gatech.edu},    
\author[rahul]{Qinsheng Zhang}\ead{qzhang419@gatech.edu},    
\author[yongxin]{Yongxin Chen}\ead{yongchen@gatech.edu}  

\address[rahul]{Machine Learning Center, Georgia Institute of Technology, Atlanta, GA, USA}  
\address[yongxin]{School of Aerospace Engineering, Georgia Institute of Technology, Atlanta, GA, USA}        



\begin{abstract}
In this paper, we propose an algorithm for estimating the parameters of a time-homogeneous hidden Markov model (HMM) from aggregate observations. This problem arises when only the population level counts of the number of individuals at each time step are available, and one seeks to learn the individual HMM from these observations. Our algorithm is built upon the classical expectation-maximization algorithm and the recently proposed aggregate inference algorithm (Sinkhorn belief propagation). 
We present the parameter learning algorithm for two different settings of HMMs: one with discrete observations and one with continuous observations, and the algorithm exhibits convergence guarantees in both cases.
 Moreover, our learning framework naturally reduces to the standard Baum-Welch learning algorithm for HMMs when the population size is $1$.  The efficacy of our algorithm is demonstrated through several numerical experiments. 
\end{abstract}

\end{frontmatter}

\section{Introduction}
\label{sec:intro}

There has been a growing interest in applications where data about individuals are not accessible, instead aggregate population-level observations in the form of counts of the individuals are available~\cite{SheDie11,LuoXuZhe16}. For various reasons including measurement fidelity, privacy preservation, cost of data collection, and scalability, data is often collected as aggregates. For example, in human ensemble flow analysis, individual trajectories may not be readily accessible due to privacy concerns, but the number of individuals in a certain geographical area can typically be counted by cell phone carriers. More examples include voter turnout based on demography from census data~\cite{Kin13} and bird migration analysis~\cite{SunSheKum15}. One fundamental part in modeling such aggregate data is estimating the individual model parameters. Learning the underlying individual model from aggregate observations is a challenging task since the full trajectory of each individual is not accessible. 

We are interested in learning hidden Markov models (HMMs) using aggregate data. HMMs are popular graphical models used in various scenarios involving unobservable (hidden) data sequences arising in ecology, social dynamics, and emergence of an epidemic~\cite{RabJua86,CapMouRyd06,DonPenHel12,SinHaaZha20}. Due to their ability to address the nonstationarity in observed data sequences, HMMs are capable of modeling a rich class of problems. In aggregate HMM settings, a large set of homogeneous individuals transit from one state to another according to the underlying HMM and at each time-step, corresponding aggregated observations are recorded. For example, in epidemiology, one can model spread of an infectious disease such as COVID-19 over time in a geographical area using the population level aggregate data generated by an HMM. In this work, we consider the problem of estimating the parameters of a time-homogeneous hidden Markov model, i.e., transition and observation probabilities, from noisy aggregate data.

A traditional method for learning HMM is the Baum-Welch algorithm~\cite{BauEag67,BauPetSou70}, which is a special case of the expectation-maximization (EM) algorithm~\cite{DemLaiRub77,NeaHin98}. For the given observations sampled from a model consisting of latent variables (variables that are not observable) with unknown parameters, the EM algorithm aims to find the maximum likelihood estimates of the model parameters. In its first step (E-step), the EM algorithm estimates a function of the expected values of the latent variables and subsequently in the second step (M-step), it finds the maximum likelihood parameter estimates. For the case of learning HMM parameters, inference algorithms such as belief propagation (BP) algorithm \cite{Pea88} is utilized in the E-step of the EM algorithm. The Baum-Welch algorithm for estimating an HMM uses the forward-backward inference algorithm, one type of BP algorithms, in the E-step to complete the data. Unfortunately, traditional HMM learning methods such as Baum-Welch algorithm~\cite{BauPetSou70} can not be applied to aggregate setting. Learning the individual model from such population-level observations becomes challenging since great amount of information about individuals is lost due to data aggregation and observation noise. 

Recently, the learning and inference problems in aggregate settings have been formalized under the collective  graphical model (CGM) framework~\cite{SheDie11}. Within the CGM framework, for learning the parameters of the individual model, several aggregate inference methods such as non-linear belief propagation (NLBP)~\cite{SunSheKum15} and Bethe-RDA~\cite{VilBelSheMcc15} algorithms has been utilized in the E-step of the EM algorithm aiming to maximize the complete data likelihood. Both of the inference algorithms work on an explicit observation model. 
In addition, since NLBP does not exhibit convergence guarantee, it does not lead to stable learning methods.


The primary contribution of our work is a novel algorithm for estimating the HMM parameters with theoretical guarantees from noisy aggregate observations. We utilize a modified EM algorithm for the learning task, where the E-step of the algorithm is solved using recently proposed aggregate inference method, the Sinkhorn belief propagation (SBP) algorithm~\cite{SinHaaZha20}. \singh{We show that our algorithm exhibits convergence guarantee.} Instead of explicitly considering the noise model, we incorporate \singh{observation} noise in the underlying graph and as a result, our algorithm reduces to the standard Baum-Welch algorithm when only one individual is considered. We further extend our algorithm to learn the model parameters with \textit{continuous} observation noise model. We evaluate the performance of our algorithm on a variety of scenarios including human ensemble flow on real-world data.

\textbf{Related Work:} Estimating Markov chains from aggregate data, also referred to as \textit{macro} data in earlier works, has a long history. It was first studied in \cite{LeeJudZel70} where the transition matrices were estimated based on maximum likelihood method. In \cite{Sun75,Mac77,KalLawVol83}, the modeling of a single Markov chain was studied by maximizing the aggregate posterior. More recent learning methods from aggregate data include~\cite{LuoXuZhe16,PasFuBou12}. \singh{After the introduction of the CGM framework in \cite{SheDie11}, there have been a few works on learning the underlying individual model from aggregate data.} The NLBP algorithm~\cite{SunSheKum15}, a message passing type algorithm for approximate inference in CGMs, has been utilized in EM for the task of learning a Markov chain. Another existing aggregate inference algorithm utilized in the E-step of the EM algorithm is Bethe-RDA~\cite{VilBelSheMcc15} which exhibits convergence guarantees.
Finally, \cite{BerShe16} proposed a method of moments estimator for learning a Markov chain within the CGM framework. Other works along this line include estimating spatiotemporal population flow~\cite{IwaShi19} and
recurrent estimation of HMM~\cite{LyuGriDun19} from aggregate data, learning stochastic behaviour of aggregate data~\cite{MaLiuShu20}, learning hidden nonlinear dynamics from aggregate data\cite{WanBoLin18}, and estimating group behavior from ensemble observations \cite{Zen19}.

The rest of the paper is organized as follows. In Section~\ref{sec:background}, we briefly discuss related background. We present our main results and algorithms in Section~\ref{sec:main_results} for discrete observations. The counterpart with continuous observations is developed in Section \ref{sec:hmm_cts} followed by experimental results in Section~\ref{sec:experiments} and a concluding remark in Section~\ref{sec:conclusion}.

\section{Background}
\label{sec:background}
In this section, we present related background on HMMs, their extension to aggregate settings, and the CFB inference algorithm. 

\subsection{Hidden Markov Models}
\label{sec:HMMs}
An HMM is a Markov chain where the variables are not directly observable, but corresponding noisy variables are observed. Denote the unobserved hidden variables as $X_1,X_2,\ldots$ and observed variables as $O_1,O_2,\ldots$. Here $X_t$ and $O_t$ are random variables taking values from sets $\cX$ and $\cO$ respectively. In general, both $\cX$ and $\cO$ can be either finite sets or infinite sets. \singh{For discrete HMMs, $\cX$ and $\cO$ are finite sets with cardinalities $|\cX| = d$ and $|\cO| = s$, respectively. }

\begin{figure}[h]
\centering
\includegraphics[scale=0.65]{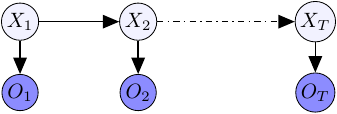}
\caption{A length $T$ HMM.}
\label{fig:hmm_graphical_model}
\end{figure}

A time-homogeneous HMM is parameterized by the initial distribution $\pi(X_1)$, the state transition probabilities $p(X_{t+1} | X_t)$, and the observation probabilities $p(O_{t} \mid X_{t})$ 
\singh{independent of time steps} $t = 1,2,\ldots$. An HMM is a special type of probabilistic graphical model (PGM) \cite{WaiJor08}. The graphical representation of a length $T$ HMM is shown in Figure~\ref{fig:hmm_graphical_model}.
The joint distribution of an HMM with length $T$ factorizes as
\begin{align} \label{eq:HMM_distribution}
    p(\bx, \bo) 
    &= \pi(x_1)~ \prod_{t=1}^{T-1} ~p(x_{t+1} \mid x_{t}) ~ \prod_{t=1}^{T}~p(o_t \mid x_t),
\end{align}
where $\bx = \{x_1,x_2,\ldots,x_T \}$ and $\bo= \{o_1,o_2,\ldots,o_T \}$ denote particular assignments to the hidden and observation variables, respectively.

One of the most important problems in HMMs is Bayesian inference where the goal is to calculate the posterior distributions of the hidden states $X_t$ given a sequence of observations $\bo= \{o_1,o_2,\ldots,o_T \}$. This is also known as filtering/smoothing~\cite{Mur12} in systems and control community.
A well-known algorithm for this task is the standard forward-backward algorithm \cite{RabJua89}, which itself a special case of belief propagation \cite{Pea88} for Bayesian inference of general graphical models. 

Another important problem in HMMs is the parameter learning, which is also known as system identification. 
Denote the set of parameters to be learned as 
	\begin{equation}\label{eq:parameterstheta}
		\theta = \{ \pi(x_1), p(x_{t+1} | x_{t}), p(o_{t} | x_{t})\}.
	\end{equation} 
Let $\{\bo^{(m)}\}_{m=1}^M$ with $\bo^{(m)}=\{o_1^{(m)},o_2^{(m)},\ldots,o_T^{(m)} \}$ be a set of observed trajectories. The objective of parameter learning of HMMs is to estimate the parameter $\theta$ using the available data $\{\bo^{(m)}\}_{m=1}^M$. Since the HMM is a latent variable model where the latent variable $X_t$ is not observable, the maximum likelihood estimation cannot be applied directly. A popular approach for learning latent variable models is the expectation-maximization (EM) algorithm \cite{NeaHin98,Bis06}. The EM algorithm is an iterative method that involves two steps in each iteration: E-step and M-step. In the E-step, the values associated with the hidden variables are estimated to make the data complete and then, in M-step, the parameters of the underlying model are optimized based on the complete data likelihood. When specialized to HMMs, the EM algorithm reduces to the Baum-Welch algorithm~\cite{KolFri09}.
\subsection{Aggregate Hidden Markov Models}
\label{subsec:collective_hmm}
Aggregate HMM is a framework for learning and inference from noisy aggregate data \singh{generated} from an HMM describing the behavior of individuals. It is a special case of the collective graphical model~\cite{SheDie11}, which is a framework for general probabilistic graphical models. The aggregate data is generated from $M$ independent individuals \singh{following an HMM}. The HMMs are aggregate in the sense that they are indistinguishable to each other.
\begin{figure}[h]
\centering
\includegraphics[scale=0.65]{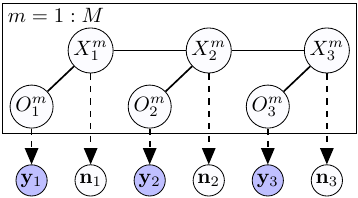}
\caption{Observation model of aggregate HMMs (shaded nodes represent aggregate observations).}
\label{fig:noise_model}
\end{figure}
Let $X_t^{(m)}$ be the (unobservable) state of the $m$-th individual at time $t$ and $O_t^{(m)}$ be the observable state. The observations are made in the form of 
$y_t(o_t) =  \sum_{m=1}^M \mathbb{I}[O_t^{(m)}= o_t] = n_t^o(o_t)$, where $\mathbb{I}$ denotes the indicator function. It is the histogram of $M$ observations over $\cO$. The aggregate observation model for length of $T=3$ is depicted in Figure~\ref{fig:noise_model}. Given these aggregate observations, the goal of inference in aggregate HMMs is to estimate the latent distributions $n_{t,t+1}(x_t,x_{t+1}) = \sum_{m=1}^M \mathbb{I}[X_t^{(m)}= x_t, X_{t+1}^{(m)}= x_{t+1}]$, $n_{t,t}(x_t,o_{t}) = \sum_{m=1}^M \mathbb{I}[X_t^{(m)}= x_t, O_t^{(m)}= o_t]$, and $n_t(x_t) = \sum_{m=1}^M \mathbb{I}[X_t^{(m)}= x_t]$. 

The exact inference is proved to be computationally infeasible~\cite{SheDie11} for problems with large $T$ and $M$. 
It is proposed in \cite{SinHaaZha20} that this aggregate inference can be approximately achieved by solving a free energy minimization problem. Moreover, the approximation error vanishes as the size $M$ of the population goes to infinity. 
 The integral constraints on $\bn$ can be relaxed \cite{SinHaaZha20} without affecting the precision much. With this relaxation, the latent distributions $\bn=\{\bn_{t},\bn_{t}^o,\bn_{t,t},\bn_{t,t+1}\}$ satisfy the local polytope constraints
\begin{subequations}\label{eq:nconstraints}
    \begin{align}
& \sum_{x \in \cX} n_{t}(x)=M, \forall t \in \{1,\cdots, T\} \label{eq:M_cons_simplex} \\
& \sum_{x \in \cX} n_{t,t+1}(x,x_{t+1})=n_{t+1}(x_{t+1}), \nonumber \\ 
& \sum_{x \in \cX} n_{t,t+1}(x_t,x)=n_t(x_t), \quad\forall t \in \{1,\cdots, T-1\}  \label{eq:cons_tt_t} \\
& \sum_{o \in \cO} n_{t,t}(x,o)=n_t(x), \quad\forall t \in \{1,\cdots, T\}                \label{eq:cons_tilde_tt_t}                   \\
         & \sum_{x \in \cX} n_{t,t}(x,o)=n_{t}^o(o), \quad\forall t \in \{1,\cdots,T\}.\label{eq:cons_tilde_tt_o}
    \end{align}
\end{subequations}
Denote the local polytope described in \eqref{eq:nconstraints} by $\mathbb{M}$~\cite{WaiJor08}. For \singh{aggregate} HMMs, the free energy equals \cite{WaiJor08,SinHaaZha20}
\begin{align} \label{eq:bethe}
    &\mathcal{F}(\bn,\theta) = 
    - \sum_{t=1}^T \sum_{x_t,o_t} n_{t,t}(x_t,o_t) \log p(o_t | x_t) \\ \nonumber
    &- \sum_{t=1}^{T-1} \sum_{x_t,x_{t+1}} {n}_{t,t}(x_t,x_{t+1}) \log p(x_{t+1} | x_t)  \\ \nonumber
    &- \sum_{x_1} n_{1}(x_1) \log \pi(x_1) -\sum_{x_1} n_1(x_1)\log n_1(x_1) 
    \\ \nonumber
    & - 2\sum_{t=2}^{T-1}\sum_{x_t} n_t(x_t) \log n_t(x_t) -\sum_{x_T}n_T(x_T)\log n_T(x_T)
    \\ \nonumber
    & + \sum_{t=1}^T \sum_{x_t,o_t} n_{t,t}(x_t,o_t) \log n_{t,t}(x_t,o_t) \\ \nonumber
    &+ \sum_{t=1}^{T-1} \sum_{x_t,x_{t+1}} {n}_{t,t+1}(x_t,x_{t+1}) \log {n}_{t,t+1}(x_t,x_{t+1}).
\end{align}
It is in fact equal to the Kullback-Leibler divergence between the inferred distribution and the prior distribution over the space of trajectories \cite{SinHaaZha20}. 
The aggregate inference problem is equivalent \cite{SinHaaZha20} to the following convex optimization problem.
\begin{prob}\label{prob:bethe}
    \begin{subequations} \label{eq:argmin-bethe}
        \begin{align}
            \min_{\bn\in \mathbb{M}} &~ \mathcal{F}(\bn, \theta)\\
            \mbox{subject to} & ~\bn_t^o = \by_t, \quad \forall t\in\{1,\cdots,T\}.
        \end{align}
    \end{subequations}
\end{prob}
Thanks to the large deviation theory \cite{Var84,SinHaaZha20}, the conditional distribution $p(\bn | \by,\theta)$ of $\bn$ given the observation $\by$ approximately concentrates on the solution to Problem \ref{prob:bethe}. The very same approximation is the foundation of the Schr\"odinger bridge problem \cite{CheGeoPav14e,CheGeoPav21} which has been explored extensively in stochastic control. 

In~\cite{SinHaaZha20}, we proposed the SBP algorithm for solving aggregate inference problems over more general CGMs with tree-structure. 
The SBP algorithm has convergence guarantees with linear rate~\cite{SinHaaZha20}.
There exist some other algorithms for aggregate inference problems in CGMs including approximate MAP \cite{SheSunKumDie13}, NLBP~\cite{SunSheKum15} and Bethe-RDA~\cite{VilBelSheMcc15}. One major difference between SBP and these methods \singh{is the observation model}. In \cite{SheSunKumDie13,SunSheKum15,VilBelSheMcc15}, the noise is added to the aggregate observation $\by_t$ directly, meaning the real observed histogram is a perturbed version of $\by_t$ by some random noise. In contrast, in Problem \ref{prob:bethe}, we assume that the observation noise enters the system in the individual level and the measurement of the histogram is precise. It has the nice property that when $M=1$, it reduces to a standard inference problem for PGMs or HMMs. We refer the reader to~\cite{SinHaaZha20} for more details on the comparison of the observation models. 
\subsection{Collective Forward-Backward Algorithm} \label{subsec:afb}
\begin{figure}[h]
\centering
\includegraphics[scale=0.6]{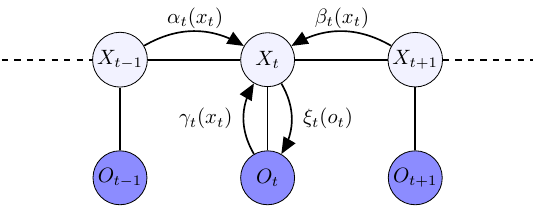}
\caption{Messages for inference in aggregate HMMs.}
\label{fig:hmm_message}
\end{figure}

The collective forward-backward algorithm (CFB) is a special case of the general SBP algorithm when the underlying graphical model is an HMM~\cite{SinHaaZha20}.
It is a message passing type algorithm, similar to BP, consisting of four types of messages. Figure~\ref{fig:hmm_message} depicts the messages employed by the CFB algorithm  with $\alpha_t(x_t)$ being the messages in the forward direction and $\beta_t(x_t)$ being the messages in the backward direction. Moreover, $\gamma_t(x_t)$ denote the messages from observation node to hidden node and $\xi_t(o_t)$ are the messages from hidden nodes to observation nodes. These messages are characterized by
\begin{subequations}\label{eq:forward_backward}
\begin{eqnarray}
    &&\alpha_t(x_t) \propto \sum_{x_{t-1}} p(x_t|x_{t-1}) \alpha_{t-1} (x_{t-1}) \gamma_{t-1}(x_{t-1}) \label{eq:forward_backward1}  \\
    &&\beta_t(x_t) \propto  \sum_{x_{t+1}} p(x_{t+1}|x_{t}) \beta_{t+1} (x_{t+1}) \gamma_{t+1}(x_{t+1}) \label{eq:forward_backward2} \\
    &&\gamma_t(x_t) \propto \sum_{o_{t}} p(o_{t}|x_{t}) \frac{y_t(o_t)}{\xi_t(o_t)} \label{eq:forward_backward3} \\
    &&\xi_t(o_t) \propto \sum_{x_{t}} p(o_{t}|x_{t}) \alpha_{t} (x_{t}) \beta_{t}(x_{t}),\label{eq:forward_backward4}
\end{eqnarray}
\end{subequations}
with boundary conditions $\alpha_1(x_1) = \pi(x_1), \beta_T(x_T) = 1$.
The sequence of update steps are listed in Algorithm~\ref{alg:collective_forward_backward}.
\begin{algorithm}[t]
   \caption{Collective Forward-Backward algorithm}
   \label{alg:collective_forward_backward}
\begin{algorithmic}
   \STATE Initialize all the messages $\alpha_t(x_t), \beta_t(x_t), \gamma_t(x_t), \xi_t(o_t)$ 
   \WHILE{not converged}
   \STATE \textbf{Forward pass:}
   \FOR{$t = 2,3,\ldots,T$}
        \STATE i) Update  $\gamma_{t-1}(x_{t-1})$ 
        \STATE ii) Update $\alpha_t(x_t), \xi_t(o_t)$
    \ENDFOR
    \STATE \textbf{Backward pass:}
    \FOR{$t = T-1,\ldots,1$}
        \STATE i) Update  $\gamma_{t+1}(x_{t+1})$ 
        \STATE ii) Update $\beta_t(x_t), \xi_t(o_t)$
    \ENDFOR
    \ENDWHILE
\end{algorithmic}
\end{algorithm}
Once the algorithm converges, which is guaranteed, the latent marginals can be estimated as
    \begin{eqnarray*}
       && n_t(x_t) \propto \alpha_t(x_t) \beta_t(x_t) \gamma_t(x_t), \label{eq:marginals_hmm_a} \\
       && n_{t,t+1}\!(x_t,\!x_{t+1})\! \propto\! p(x_{t+1} | x_t)  \alpha_t(x_t) \gamma_t(x_t) \beta_t(x_{t+1})  \gamma_t(x_{t+1}) \label{eq:marginals_hmm_b}  \\
       && n_{t,t}(x_t,o_t) \propto \frac{p(o_t | x_t) \alpha_t(x_t) \beta_t(x_t)}{\xi_t(o_t)}.
       \label{eq:marginals_hmm_c}  
    \end{eqnarray*}
Clearly, when the population size is $1$, i.e., $M=1$, the CFB algorithm reduces to the standard Forward-backward algorithm for the inference of HMMs \cite{SinHaaZha20,HaaSinZha2020}. 

\section{Learning Discrete Aggregate HMMs}
\label{sec:main_results}
The learning problem in CGMs is concerned with estimating the individual model parameters of the underlying graphical model from aggregate observations. For learning the parameters of a latent variable model, the EM algorithm~\cite{DemLaiRub77} is a standard approach. The EM algorithm consists of two operations: the E-step to compute the log-likelihood of the observations given the current estimation of parameters, and the M-step to maximize the log-likelihood. The challenge to apply the EM algorithm for learning CGMs lies in the fact that the E-step requires inferring the conditional distribution of $\bn$ on the observation $\by$, which is untractable \cite{SheSunKumDie13,SinHaaZha20}. 

In this section, we propose the approximate EM algorithm (Algorithm \ref{alg:mm_cgm}) for learning HMMs with observations in aggregate form. The key idea is to use the tractable CFB algorithm to approximately infer the aggregate distributions $\bn$. Note that the SBP algorithm can be used for learning more general CGMs. 
\begin{algorithm}[t]
    \caption{Approximate EM algorithm}
    \label{alg:mm_cgm}
 \begin{algorithmic}
    \STATE Initialize model parameters $\theta^{0}$
    \FOR{$\ell = 1,2,\ldots$} 
    \STATE E-step: Obtain the solution $\bn^*$ to Problem \ref{prob:bethe} using CFB with parameters $\theta^{\ell-1}$
    \STATE M-step: $\theta^{\ell} = \argmin_{\theta} \cF(\bn^*,\theta)$
     \ENDFOR
 \end{algorithmic}
\end{algorithm}
%
\begin{thm}
    \label{thm:mean-ll}
The Approximate EM algorithm converges.
\end{thm}
\begin{proof}
The E-step and M-step in Algorithm \ref{alg:mm_cgm} are coordinate descent updates of the free energy $\cF(\bn,\theta)$  with respect to $\bn$ and $\theta$, and thus the objective function is monotonically decreasing. Moreover, since the free energy $\cF$ in \eqref{eq:bethe} is equal to Kullback-Leibler divergence between the inferred distribution and the prior distribution over the space of trajectories \cite{SinHaaZha20}, it is bounded below by $0$. Thus, in view of the fact that $\cF$ is continuously differentiable, the approximate EM algorithm converges to a local minimum. 
\end{proof}

We next argue that the log-likelihood $L(\theta^{\ell}):=\log p(\by|\theta^{\ell})$ approximately monotonically increases.
The improvement of $L$ at the $\ell$-th iteration is
\begin{align*}
& L(\theta^\ell) \!- L(\theta^{\ell-1}) 
\!= \mathrm{log}\sum_{\bn} p(\by, \bn \mid \theta^\ell) \!- \mathrm{log}~p(\by \mid \theta^{\ell-1}) \\
&~ = \mathrm{log}~ \sum_{\bn} p(\bn | \by , \theta^{\ell-1}) \frac{p(\by, \bn \mid \theta^\ell)}{p(\bn | \by , \theta^{\ell-1})}  - \mathrm{log}~p(\by \mid \theta^{\ell-1}) \\
&~ \geq  \sum_{\bn} p(\bn | \by , \theta^{\ell-1}) ~\mathrm{log}~\frac{p(\by, \bn \mid \theta^\ell)}{p(\bn | \by , \theta^{\ell-1})}  - \mathrm{log}~p(\by \mid \theta^{\ell-1}) \\
&~ =  \sum_{\bn} p(\bn | \by , \theta^{\ell-1}) ~\mathrm{log}~\frac{p(\by, \bn \mid \theta^\ell)}{p(\by, \bn \mid  \theta^{\ell-1})}.
\end{align*}
Since $p(\bn | \by,\theta^{\ell-1})$ approximately concentrates on $\bn^*$,
\begin{align*}
    L(\theta^\ell) - L(\theta^{\ell-1}) &\approx \log\frac{p(\by,\bn^* \mid \theta^\ell)}{p(\by,\bn^* \mid \theta^{\ell-1})}.
\end{align*}
Again, due to the large deviation theory \cite{SinHaaZha20}, $p(\by,\bn \mid \theta)\approx \exp[-M\cF(\bn,\theta)]$. Thus,
\[
    \frac{1}{M}(L(\theta^\ell) - L(\theta^{\ell-1}))  \approx  -\cF(\bn^*, \theta^\ell)+ 
    \cF(\bn^*, \theta^{\ell-1}).
\]
The approximate monotonicity of likelihood then follows from the definition of the M-step in Algorithm \ref{alg:mm_cgm}.



Thanks to the special structure of HMMs, the M-step can be implemented efficiently in closed-form.
\begin{prop}
    \label{prop:Maximization_HMM}
    The M-step in learning aggregate HMMs is given by
    \begin{subequations}\label{eq:parameters}
    \begin{align}
         \pi(x_1) &= n_1(x_1),  \label{eq:parameters_a} \\
        p(x_{t+1} \mid x_t) &= \frac{\sum_{t=1}^{T-1} n_{t,t+1}(x_t,x_{t+1})}{\sum_{t=1}^{T-1} n_{t}(x_t)}, \label{eq:parameters_b}  \\
        p(o_{t} \mid x_t) &= \frac{\sum_{t=1}^{T} n_{t,t}(x_t,o_{t})}{\sum_{t=1}^{T} n_{t}(x_t)}. \label{eq:parameters_c}  
    \end{align}
    \end{subequations}
\end{prop}
\begin{proof}
See Appendix~\ref{appendix:proof_thm3}.
\end{proof}
\begin{rem}
If parts of the parameters are known, then we only need to update the other parameters in the M-step. For instance, if the emission probability $p(o_t|x_t)$ is known, then only \eqref{eq:parameters_a}-\eqref{eq:parameters_b} are needed for the M-step.
\end{rem}


Algorithm~\ref{alg:mm_cgm} is for learning from a single sequence of aggregate data generated from a certain number of samples.
Learning from multiple sequences of observations was initially explored in the Baum-Welch algorithm \cite{RabJua89}. Building on the same idea, we extend it to the setting (Algorithm~\ref{alg:em_hmm_discrete_ensemble}) with an ensemble of $K$ number of aggregate observation sequences generated from the same HMM model. Note that here each aggregate observation is based on the collective information of $M$ individuals, therefore $K$ such aggregate observations in fact corresponds to $N= MK$ individuals.

Denote the ensemble of aggregate observations by $\{\by^k\}_{k=1}^K$, then in the E-step, we need to find the solution $\bn^k$ to Problem \ref{prob:bethe} for each of these observations. In the M-step, one solves 
	\[
		\min_\theta \sum_{k=1}^K \cF(\bn^k, \theta).
	\]
This can again be expressed in closed form for HMMs as
	\begin{subequations}\label{eq:parameterupdateE}
	\begin{eqnarray}
		&&\pi(x_1) = \frac{1}{K}\sum_{k=1}^K n_1^k(x_1)
		\\
		&&p(x_{t+1} \mid x_t) = \frac{\sum_{k=1}^{K} \sum_{t=1}^{T-1} n_{t,t+1}^k(x_t,x_{t+1})}{\sum_{k=1}^{K} \sum_{t=1}^{T-1} n_{t}^k(x_t)}
		\\
		&&p(o_{t} \mid x_t) = \frac{\sum_{k=1}^{K} \sum_{t=1}^{T} n_{t,t}^k(x_t,o_{t})}{\sum_{k=1}^{K} \sum_{t=1}^{T} n_{t}^k(x_t)}.
	\end{eqnarray}
	\end{subequations}
\begin{algorithm}[ht]
    \caption{Learning HMMs with an ensemble of aggregate observations}
    \label{alg:em_hmm_discrete_ensemble}
 \begin{algorithmic}
    \STATE Initialize $\pi(x_1)$, $p(x_{t+1} \mid x_t)$, $p(o_{t} \mid x_t)$
    \REPEAT 
    \STATE Compute $\bn^k$ by solving Problem \ref{prob:bethe} with measurement $\by^k$ using CFB for all $k=1,\ldots,K$
    \STATE Update the parameters using \eqref{eq:parameterupdateE}
     \UNTIL{convergence}
 \end{algorithmic}
 \end{algorithm}
\begin{prop}\label{prop:connection_Baum_Welch}
Algorithms~\ref{alg:mm_cgm} and \ref{alg:em_hmm_discrete_ensemble} reduce to the Baum-Welch algorithm when observations are from populations of size $M=1$.
\end{prop}
\begin{proof}
See Appendix~\ref{appendix:proof_prop_connection_Baum_Welch}.
\end{proof}



\section{Learning Aggregate HMMs with Continuous Observations}
\label{sec:hmm_cts}
Next we turn our attention to the parameter learning problems of HMMs with continuous observation space $\cO=\mathbb{R}^s$ (the state space $\cX$ is still discrete). Such a HMM with continuous observation is similar to the discrete HMM except that it has a continuous emission density.The continuous observation model in standard HMMs has been studied in~\cite{Jua85,JuaLevSon86}. In this section, we extend our learning algorithm to aggregate HMMs with continuous emission densities. 

Suppose we have a total of $M$ trajectories of continuous observations $\{ o_1^{(m)}, o_2^{(m)}, \ldots , o_T^{(m)}\},\forall m=1,2,\ldots,M, o_t^{(m)} \in \mathbb{R}^s$ over an HMM of length $T$. 
Note that the individuals are indistinguishable, which implies the order $\{o_t^{(1)}, o_t^{(2)}, \cdots, o_t^{(M)}\}$ at each time point $t$ is arbitrary and meaningless. In discrete aggregate HMMs, the observation at time $t$ can be summarized in a histogram $\by_t$. This is not an efficient representation of observation in the setting with continuous observations as it requires discretizing the observation space $\cO$ which would be potentially expensive. Instead, we keep the observation at time $t$ in its raw format $\{o_t^{(1)}, o_t^{(2)}, \cdots, o_t^{(M)}\}$, as a bunch of samples. Similarly, since the observation space is continuous, the joint distribution $n_{t,t}(x_t,o_t)$ is no longer an efficient representation. 
We instead use $n_{t}^{(m)}(x_t)$ to capture the association between the states and the observations.

Recently, the inference problem in aggregate HMMs with continuous emission densities has been studied in~\cite{ZhaSinChe20}. It was shown that the latent marginals can be estimated as (Corollary 2, \cite{ZhaSinChe20})
\begin{subequations}\label{eq:marginals_hmm_cts}
    \begin{eqnarray}
        && n_t(x_t) \propto \alpha_t(x_t) \beta_t(x_t) \gamma_t(x_t), \label{eq:marginals_hmm_a_cts} \\
       && n_{t,t+1}(x_t,x_{t+1}) \propto p(x_{t+1} | x_t)  \alpha_t(x_t) \gamma_t(x_t) \nonumber \\
       && \qquad \qquad \qquad \qquad \qquad \beta_t(x_{t+1})  \gamma_t(x_{t+1}) \label{eq:marginals_hmm_b_cts}  \\
       && n_{t}^{(m)}(x_t)\propto \frac{p(o_t^{(m)} | x_t) \alpha_t(x_t) \beta_t(x_t)}{\xi_t(m)},
       \label{eq:marginals_hmm_c_cts}  
    \end{eqnarray}
\end{subequations}
where $\alpha_t(x_t), \beta_t(x_t),$ and $\gamma_t(x_t)$ are the messages in aggregate HMMs as depicted in Figure~\ref{fig:hmm_message}. They correspond to the fixed point of the updates
\begin{subequations}\label{eq:forward_backward_cts}
\begin{eqnarray}
    &&\alpha_t(x_t) = \sum_{x_{t-1}} p(x_t|x_{t-1}) \alpha_{t-1} (x_{t-1}) \gamma_{t-1}(x_{t-1}), \label{eq:forward_backward_cts1}  \\
    &&\beta_t(x_t) = \sum_{x_{t+1}} p(x_{t+1}|x_{t}) \beta_{t+1} (x_{t+1}) \gamma_{t+1}(x_{t+1}), \label{eq:forward_backward_cts2} \\
    &&\gamma_t(x_t) = \frac{1}{M} \sum_{m=1}^{M} \frac{p(o_{t}^{(m)}|x_{t})} {\xi_t(m)}, \label{eq:forward_backward_cts3} \\
    &&\xi_t(m) = \sum_{x_{t}} p(o_{t}^{(m)}|x_{t}) \alpha_{t} (x_{t}) \beta_{t}(x_{t}) \label{eq:forward_backward_cts4}
\end{eqnarray}
\end{subequations}
with $\alpha_1(x_1) = \pi(x_1), \beta_T(x_T) = 1$.

The inference estimates given by \eqref{eq:marginals_hmm_cts} are applicable to aggregate HMMs with any general continuous emission density. Next, we derive the formulas for parameter estimation of the underlying continuous observation HMM with Gaussian emission density.

Assuming the Gaussian noise model for emission density, it takes the form
\begin{equation}\label{eq:Gaussian}
    p(o_t|x_t) = \mathcal{N}(o_t; \mu(x_t), \Sigma(x_t)),
\end{equation}
i.e., each (discrete) hidden state corresponds to a single Gaussian density parameterized by mean $\mu(x_t)$ and variance $\Sigma(x_t)$. In such a model, an observation $o_t^{(m)}$ corresponding to the $m$-th individual at time $t$ is nothing but a sample from one of the Gaussian components. 
\begin{algorithm}[t]
    \caption{Learning aggregate Gaussian-HMMs}
    \label{alg:em_hmm_cts}
 \begin{algorithmic}
    \STATE Initialize $\pi(x_1)$, $p(x_{t+1} \mid x_t)$, $\mu(x_{t})$, $\Sigma(x_t)$
    \REPEAT 
    \STATE Compute $n_{t,t+1}(x_t,x_{t+1}),n_t(x_t),n_{t}^{(m)}(x_t)$ using CFB
    \STATE Update the parameters using \eqref{eq:parameters_cts}
     \UNTIL{convergence}
 \end{algorithmic}
 \end{algorithm}



The learning of aggregate HMMs with Gaussian emission density can be achieved using the approximate EM algorithm with slightly modifications in the two steps.
The E-step is an inference step using~\eqref{eq:marginals_hmm_cts}. The M-step has a closed-form expression given by the following Proposition. 
\begin{prop}
    \label{prop:Maximization_HMM_cts}
    The M-step for aggregate HMMs with Gaussian emission density take the form
    \begin{subequations}\label{eq:parameters_cts}
    \begin{eqnarray}
        && \pi(x_1) = n_1(x_1),  \label{eq:parameters_cts_a} \\
       && p(x_{t+1} \mid x_t) = \frac{\sum_{t=1}^{T-1} n_{t,t+1}(x_t,x_{t+1})}{\sum_{t=1}^{T-1} n_{t}(x_t)}, \label{eq:parameters_cts_b}  \\
       && \mu(x_t) = \frac{\sum_{t=1}^{T} \sum_{m=1}^M n_{t}^{(m)}(x_t)~  o_t^{(m)}} { \sum_{t=1}^{T} n_{t}(x_t) }, \label{eq:parameters_cts_c}  \\
       && \Sigma(x_t\!)\! =\! \frac{\sum_{t=1}^{T}\!\!\sum_{m=1}^M \!n_{t}^{(m)}\!(x_t)\!( o_t^{(m)}\!\!\! -\! \mu_t)\!( o_t^{(m)}\!\!\! -\! \mu_t)' }{ \sum_{t=1}^{T} n_{t}(x_t) } \label{eq:parameters_cts_d}  
    \end{eqnarray}
    \end{subequations}
    where prime denotes matrix transpose.
\end{prop}
\begin{proof}
See Appendix~\ref{appendix:proof_prop_cts}.
\end{proof}

Based on Proposition~\ref{prop:Maximization_HMM_cts}, the parameters of a Gaussian-HMM are estimated using Algorithm~\ref{alg:em_hmm_cts}. Note that in this aggregate Gaussian-HMM setting, the estimation updates for the initial distribution $\pi(x_1)$ and the transition probabilities $p(x_{t+1}|x_t)$ are the same as in Algorithm~\ref{alg:mm_cgm}. 
 \begin{rem}
 The convergence of Algorithm~\ref{alg:em_hmm_cts} follows from the same arguments as in the proof of Theorem~\ref{thm:mean-ll}.
 \end{rem}
 \begin{rem}
 Similar to discrete HMMs, one can extend Algorithm \ref{alg:em_hmm_cts} to the setting with an ensemble of continuous aggregate observations.
 \end{rem}
\begin{figure*}
    \centering
    \begin{subfigure}{.32\textwidth}
    \centering
    \includegraphics[scale=0.27]{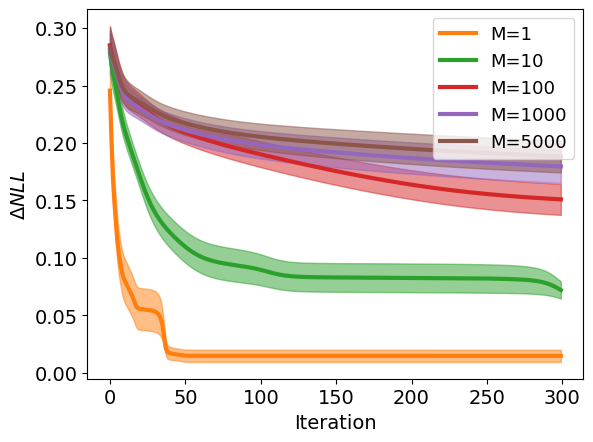}
    \caption{$d=3$}
    \label{fig:synthetic_d3}
    \end{subfigure}
    \begin{subfigure}{.32\textwidth}
    \centering
    \includegraphics[scale=0.27]{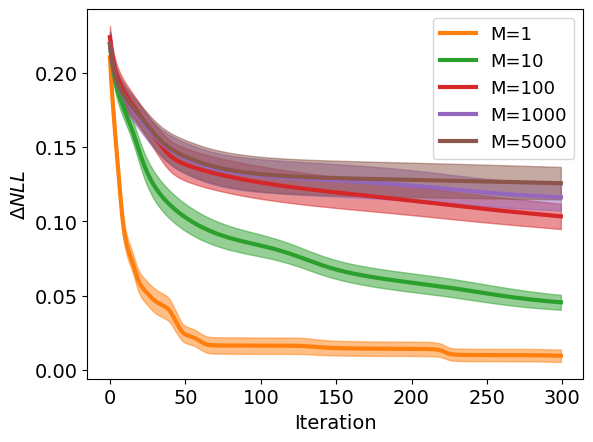}
    \caption{$d=5$}
    \label{fig:synthetic_d5}
    \end{subfigure}
    \begin{subfigure}{.32\textwidth}
    \centering
    \includegraphics[scale=0.27]{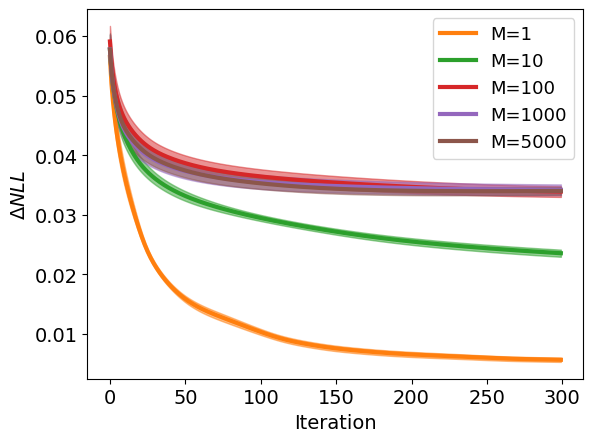}
    \caption{$d=10$}
    \label{fig:synthetic_d10}
    \end{subfigure}
    \caption{\singh{Learning} curves of HMMs with discrete observations. Curves in different color depict the results with different $M$.  All three experiments share the same values of $T=5, N=5000$. The figures show how $\Delta NLL$ evolves with the number of iterations, for $d=3,~d=5$ and $d=10$ respectively. The shaded region represents standard deviation over 10 random seeds.}
    \label{fig:synthetic}
\end{figure*}
\begin{figure*}
    \centering
    \begin{subfigure}{.27\textwidth}
    \centering
    \includegraphics[scale=0.27]{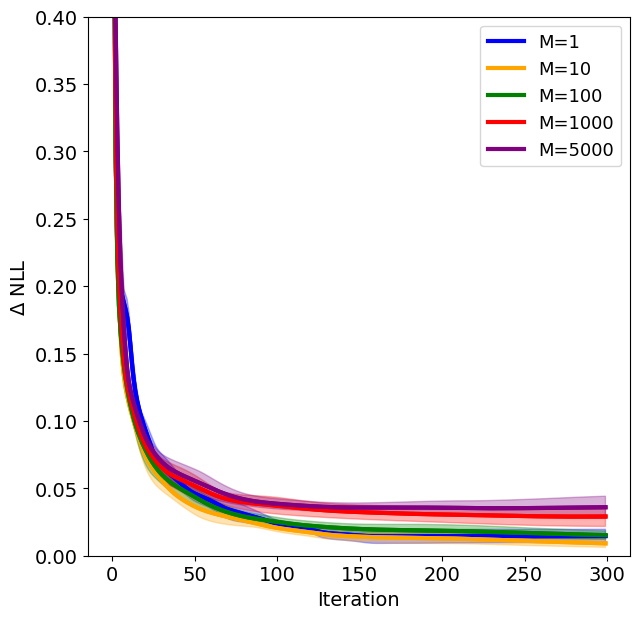}
    \caption{$d=10$}
    \label{fig:synthetic_go_d3}
    \end{subfigure}
    \begin{subfigure}{.27\textwidth}
    \centering
    \includegraphics[scale=0.27]{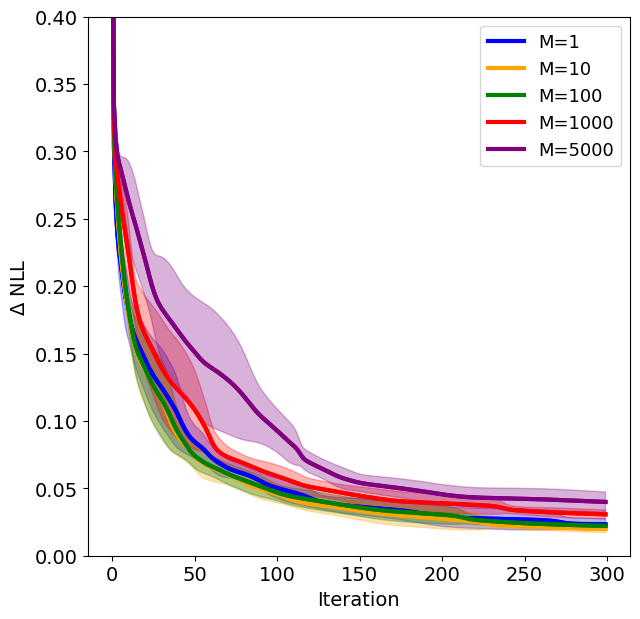}
    \caption{$d=20$}
    \label{fig:synthetic_go_d5}
    \end{subfigure}
    \begin{subfigure}{.27\textwidth}
    \centering
    \includegraphics[scale=0.27]{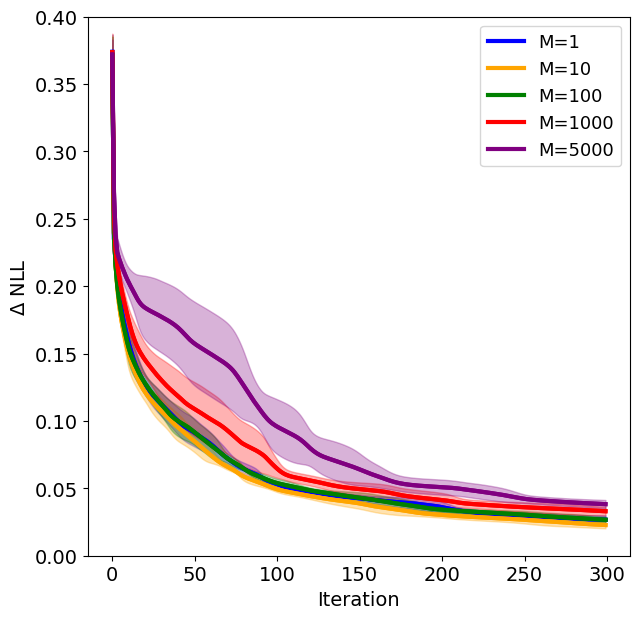}
    \caption{$d=25$}
    \label{fig:synthetic_go_d100}
    \end{subfigure}
    \caption{\singh{Learning} curves of various HMMs with Gaussian observation models. Curves in different color depict the results with different $M$. All three experiments are HMMs with $T=5$ and $N=5000$.} 
    \label{fig:synthetic_go}
\end{figure*}
\section{Experiments}
\label{sec:experiments}
To illustrate the efficacy of the proposed aggregate learning algorithms, we perform multiple sets of experiments on synthetic as well as real-world dataset of population flow over a geographical area. 

\subsection{Learning HMMs with Synthetic Data}
In this section, we consider synthetic data for evaluating our learning algorithms. We perform multiple sets of experiments for performance comparison of fitted time-invariant HMM models with discrete as well as continuous observations. The initial state probability $\pi(x_1)$ is sampled from the uniform distribution over the probability simplex.
To produce the transition matrix, we first randomly permute rows of noised identity matrix $\mathcal{I} + 0.05 \times \sqrt{d} \times \exp(Uniform[-1,1])$. We scale rows of the permuted matrix so that the resulting matrix is a valid conditional distribution. For discrete observation setting, the emission matrix is generated in a similar way as transition matrix, but with a different random seed.
In case of HMMs with continuous observations, we consider the Gaussian emission model. For each state, the corresponding Gaussian distribution is parameterized by a random mean and variance. The mean is sampled from $Uniform[-5d,5d]$ and variance is from $Uniform[1,5]$. In continuous observation setting, the algorithm is required to estimate the initial distribution, the transition matrix and the means of Gaussian emission densities.
We generate $N$ individual trajectories from the HMM parameterized with $\theta^*$ and aggregate them. Each aggregate sequence consists of collective observations of $M$ independent trajectories of length $T$. The HMM parameters are learned based on $K = \frac{N}{M}$ number of aggregate sequences. For testing purpose, we generate another set of $N$ individual trajectories.

We use the negative log likelihood ($NLL$) as a metric for evaluating performance of our learning algorithm. The difference of $NLLs$ between the learned model $\theta$ and ground truth $\theta^*$ is
\[
\Delta NLL(\theta) = \frac{1}{N} NLL(\theta) - \frac{1}{N} NLL(\theta^*).
\]
The HMM model with learned parameters is evaluated on test dataset with the same number of total trajectories $N$ as in training data.
\begin{figure*}
    \centering
    \begin{subfigure}{.27\textwidth}
    \centering
    \includegraphics[scale=0.27]{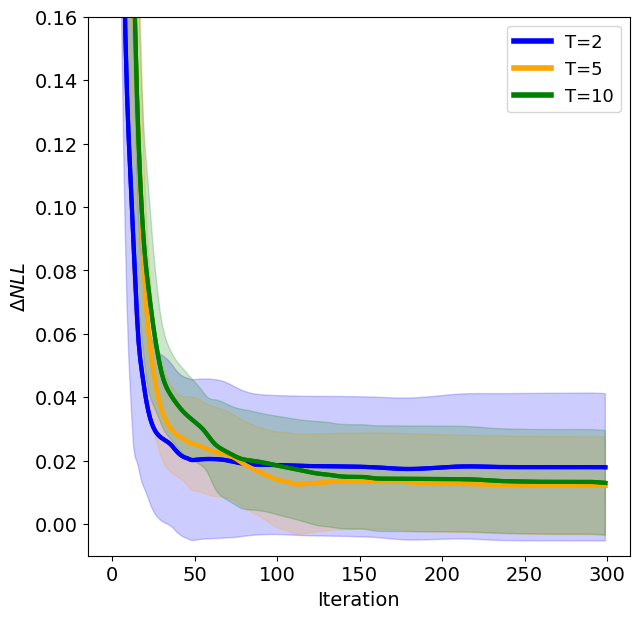}
    \caption{$M=1$}
    \label{fig:synthetic_go_T_M1}
    \end{subfigure}
    \begin{subfigure}{.27\textwidth}
    \centering
    \includegraphics[scale=0.27]{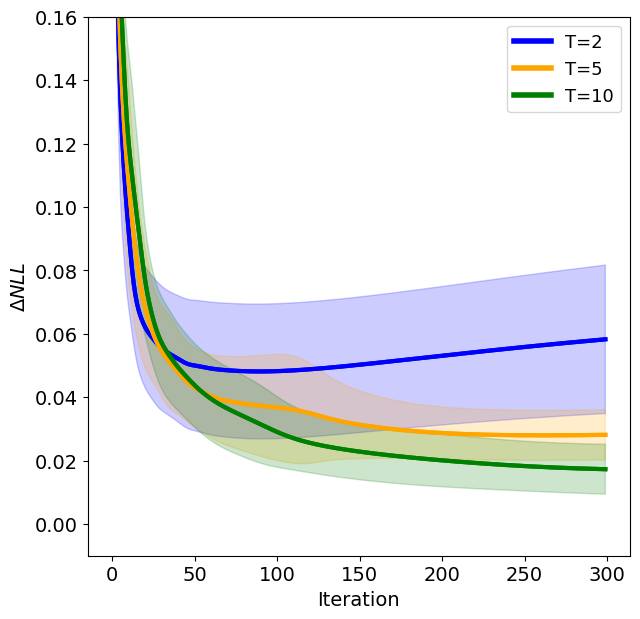}
    \caption{$M=100$}
    \label{fig:synthetic_go_T_M10}
    \end{subfigure}
    \begin{subfigure}{.27\textwidth}
    \centering
    \includegraphics[scale=0.27]{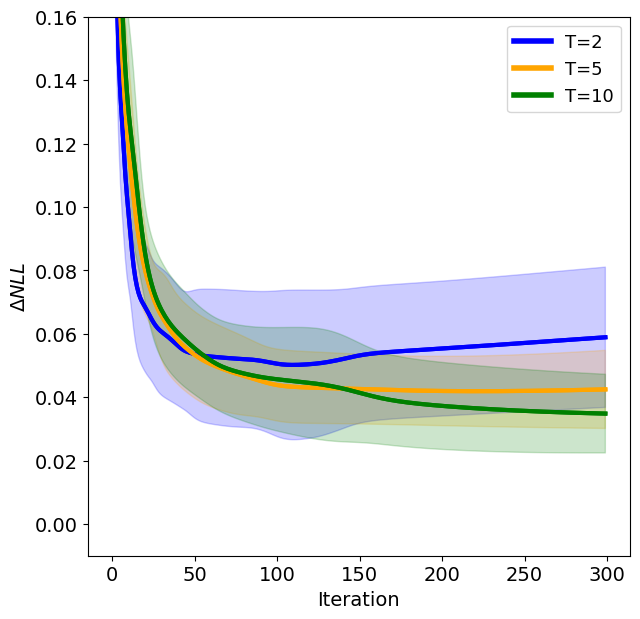}
    \caption{$M=250$}
    \label{fig:synthetic_go_T_M10}
    \end{subfigure}
    \caption{ Effect of the HMM length on the learning performance. Curves with different color correspond to different $T$ values. All three experiments are Gaussian observation HMMs with $d=5$ and $N=500$. }
    \label{fig:synthetic_go_T}
\end{figure*}
\begin{figure*}
    \centering
    \begin{subfigure}{.32\textwidth}
    \centering
    \includegraphics[scale=0.27]{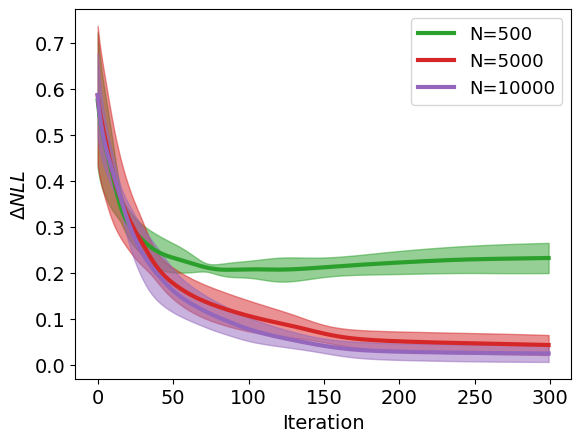}
    \caption{$M=1$}
    \label{fig:synthetic_go_traj_M1}
    \end{subfigure}
    \begin{subfigure}{.32\textwidth}
    \centering
    \includegraphics[scale=0.27]{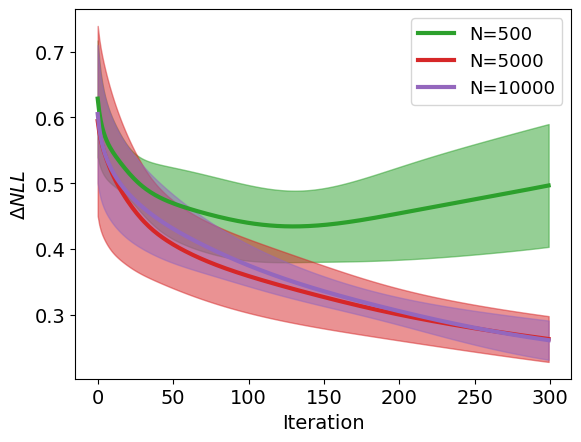}
    \caption{$M=10$}
    \label{fig:synthetic_go_traj_M10}
    \end{subfigure}
    \begin{subfigure}{.32\textwidth}
    \centering
    \includegraphics[scale=0.27]{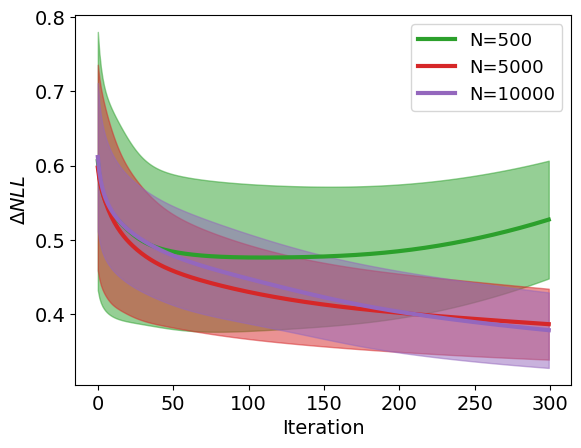}
    \caption{$M=100$}
    \label{fig:synthetic_go_traj_M100}
    \end{subfigure}
    \caption{Performance of aggregate learning with various data sizes. Curves with different color depict the learning curves with different data sizes $N$. The insufficient data causes overfitting to the training data. Our algorithm shows better performance with more samples available. All three experiments are discrete observation HMMs with $d=10,$ and $T=10$. }
    \label{fig:synthetic_go_traj}
\end{figure*}
Figure \ref{fig:synthetic} shows the performance of our algorithm for different values of state dimension $d$ and population size $M$ on HMMs with discrete observations. Curves in the same figure show learning performance with different values of $M$ but with fixed values for $d,~T$, and $N$. It can be observed that one achieves best performance for the case of no aggregation ($M=1$) and as the aggregate size $M$ increases, $\Delta NLL$ also increases. Similar observations can be made for the case of Gaussian observation model as depicted in Figure \ref{fig:synthetic_go}. It shows that our algorithm can effectively learn the generative models. Larger aggregate size corresponds to lower convergence rate as expected; with larger aggregate size, more information is lost about the individuals. 

\begin{figure*}
    \centering
    \begin{subfigure}{.32\textwidth}
    \centering
    \includegraphics[scale=0.3]{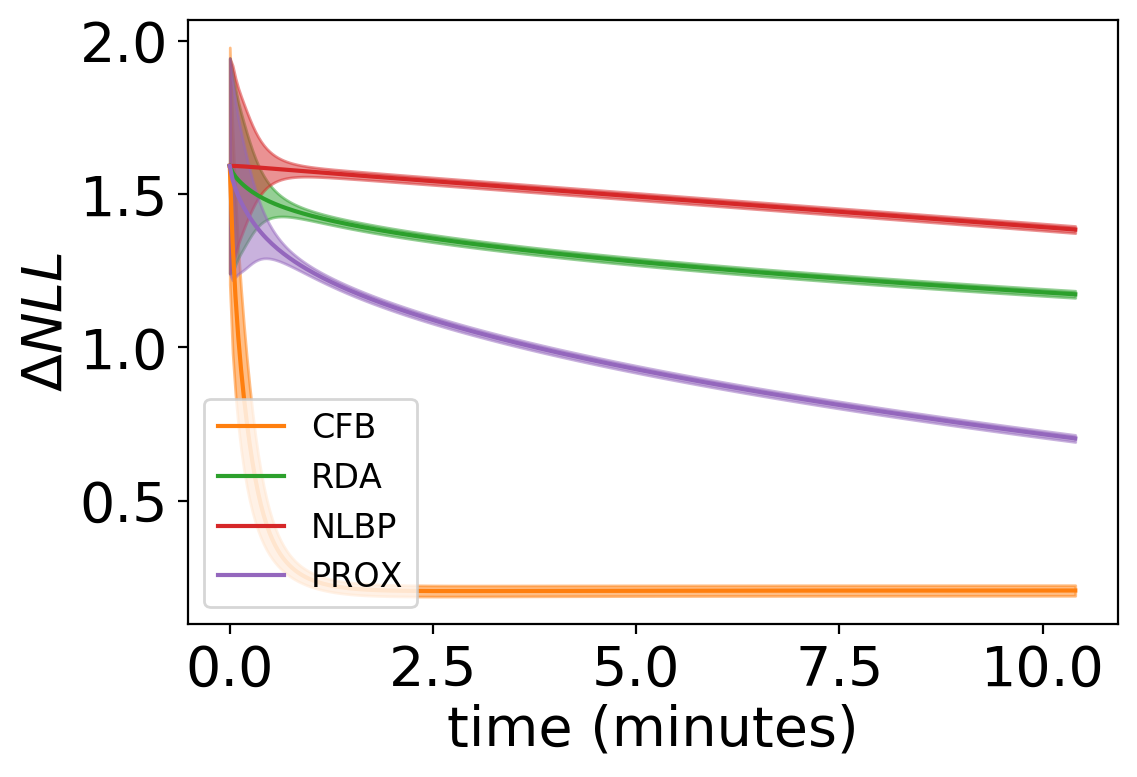}
    \caption{$d=3$}
    \label{fig:mthod_d3}
    \end{subfigure}
    \begin{subfigure}{.32\textwidth}
    \centering
    \includegraphics[scale=0.3]{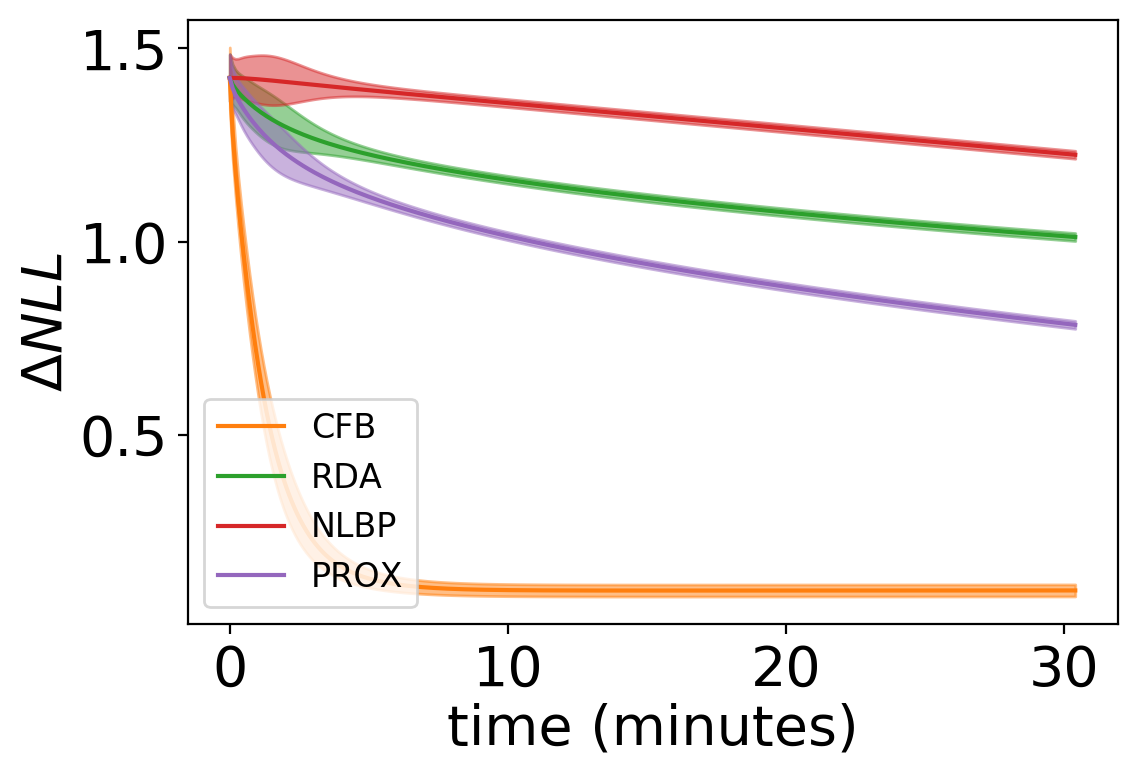}
    \caption{$d=5$}
    \label{fig:mthod_d5}
    \end{subfigure}
    \begin{subfigure}{.32\textwidth}
    \centering
    \includegraphics[scale=0.3]{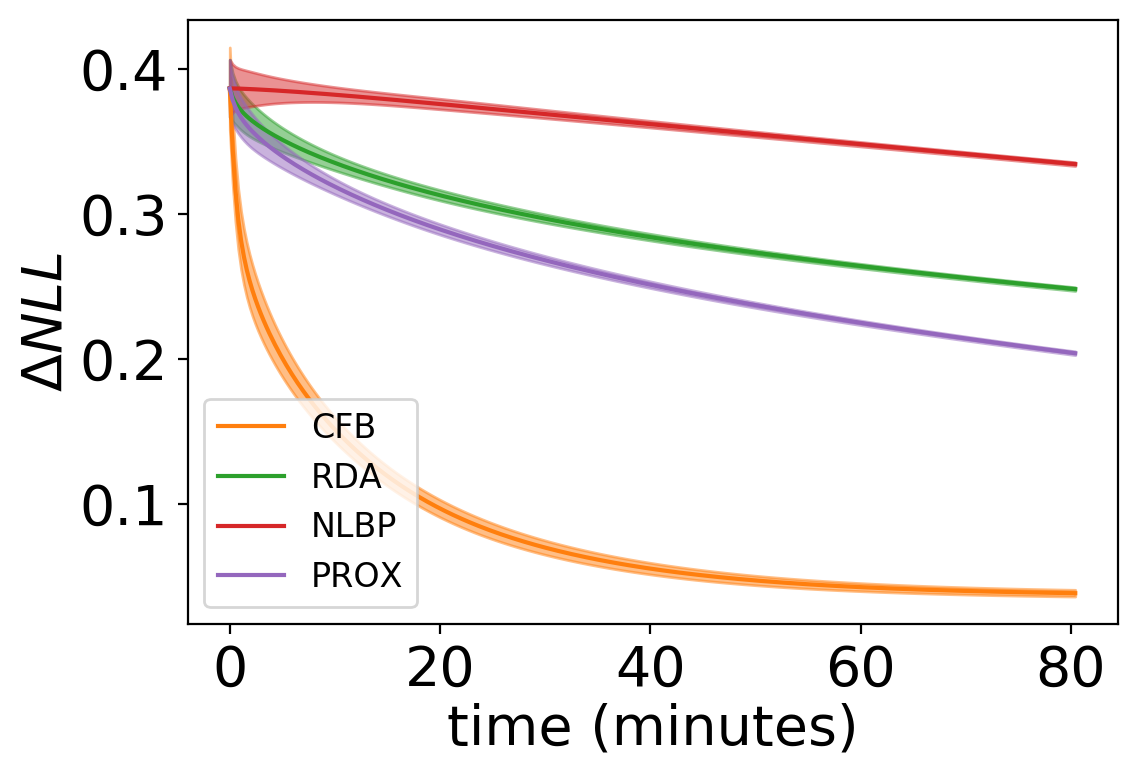}
    \caption{$d=10$}
    \label{fig:mthod_d10}
    \end{subfigure}
    \caption{Comparison of different learning algorithms under discrete HMM with $T=5,N=5000,M=10$. We report evolution  of $\Delta NLL$ with respect to wall time. It clearly shows that learning with CFB outperforms other existing approaches.}
    \label{fig:methods_cmp}
\end{figure*}
To further demonstrate the scalability of our algorithm, we conduct experiments with various HMM lengths and sample sizes as depicted in Figure \ref{fig:synthetic_go_T} and Figure \ref{fig:synthetic_go_traj}, respectively.
In Figure \ref{fig:synthetic_go_T}, the curves in different colors depict the learning performance with different HMM lengths $T$. We observe that larger $T$ leads to better performance. This is because larger $T$ is associated with more training data. Moreover, one can also observe that as the aggregate size $M$ increases, the performance degrades as expected. Figure \ref{fig:synthetic_go_traj} demonstrates the effect on HMM learning varying data sizes $N$. It can be observed that with more data available, the performance of our algorithm improves. With data size smaller than $N=500$, the overfitting problem occurs; even though the algorithm converges on training data, the $\Delta NLL$ evaluated on test data tends to increase.

Next, we compare our algorithm with the learning framework involving NLBP~\cite{SunSheKum15}, Bethe-RDA~\cite{VilBelSheMcc15}, and Prox~\cite{SinHaaZha20} in Figure~\ref{fig:methods_cmp}. Since those algorithms assume different observation models, we only learn the initial distribution and transition matrix with given observations models for a fair comparison. For the cases of NLBP, Bethe-RDA, and Prox, we choose the explicit aggregate noise model following independent Poisson distributions for each aggregate state
(see \cite{SunSheKum15} for more details on this aggregate noise model). \singh{We conduct experiments on discrete HMM with $T=5, N=5000$, and $M=10$. The comparison of learning performances for different values of $d$ is depicted in Figure~\ref{fig:methods_cmp}.} One can clearly observe from Figure~\ref{fig:methods_cmp} that our learning framework based on the CFB algorithm converges faster and performs better than the existing aggregate learning approaches.

\begin{figure}
    \centering
    \begin{subfigure}{.22\textwidth}
    \centering
    \includegraphics[scale=0.27]{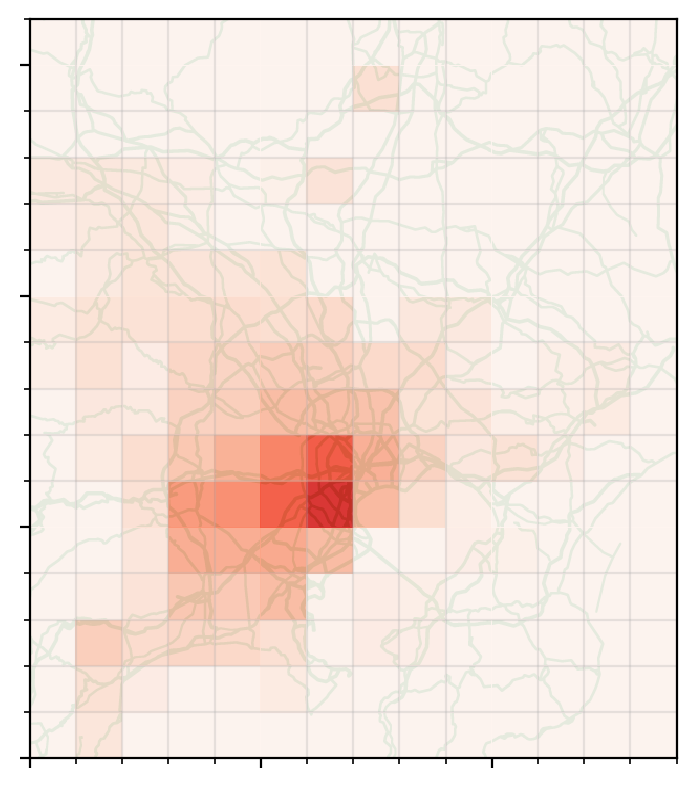}
    \caption{$2:00$}
    \label{fig:tokyo2_heat}
    \end{subfigure}
    \begin{subfigure}{.22\textwidth}
    \centering
    \includegraphics[scale=0.27]{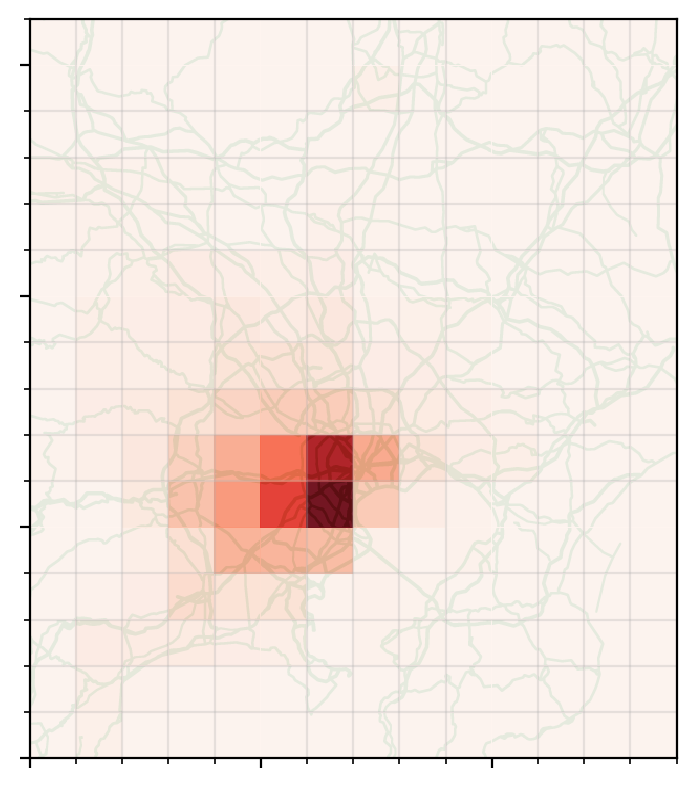}
    \caption{$14:00$}
    \label{fig:tokyo14_heat}
    \end{subfigure}
    \caption{Heatmap observation of population around the city of Tokyo. The whole area of the city is divided into $14\times16$ blocks. With more people stay in a block, color inside the block becomes deeper. The underlying green curves represent main roads around the city.}
    \label{fig:city_heat}
\end{figure}

\begin{figure}
    \centering
    \begin{subfigure}{.2\textwidth}
    \centering
    \includegraphics[scale=0.27]{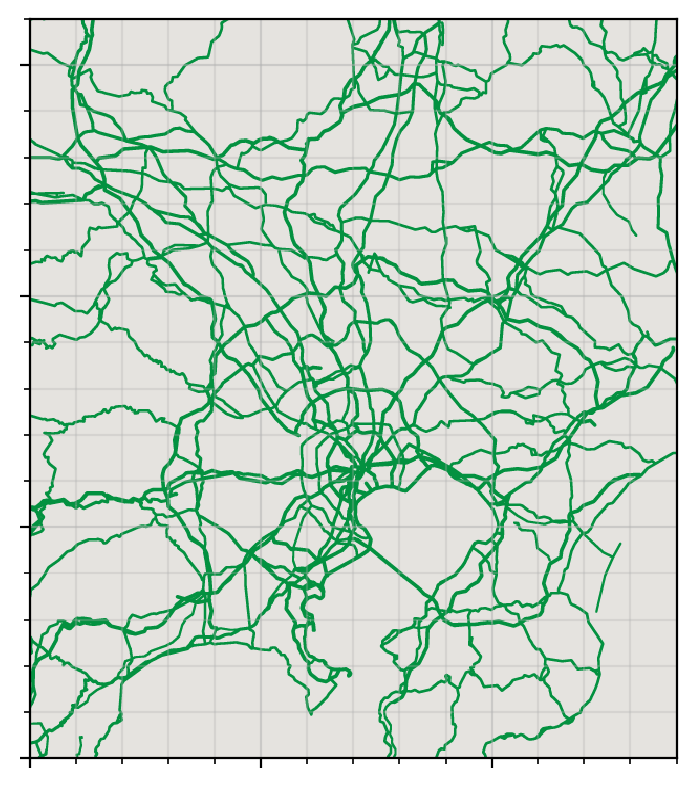}
    \end{subfigure}
    \begin{subfigure}{.2\textwidth}
    \centering
    \includegraphics[scale=0.27]{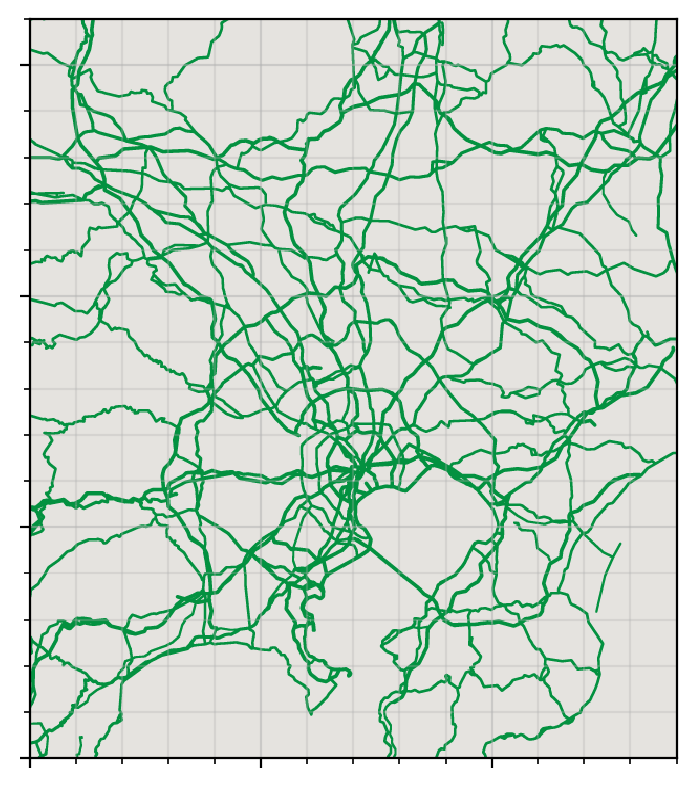}
    \end{subfigure}
    
    \begin{subfigure}{.2\textwidth}
    \centering
    \includegraphics[scale=0.27]{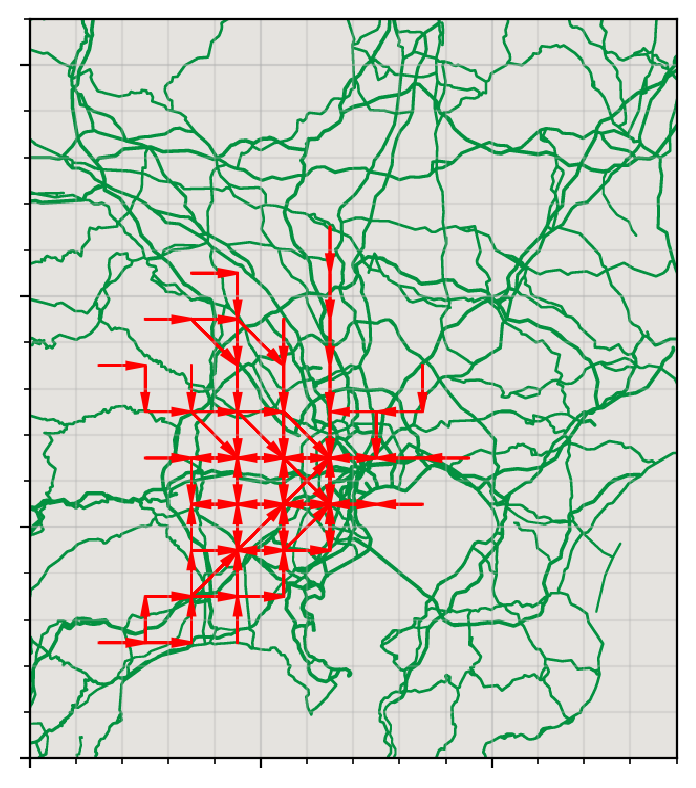}
    \end{subfigure}
    \begin{subfigure}{.2\textwidth}
    \centering
    \includegraphics[scale=0.27]{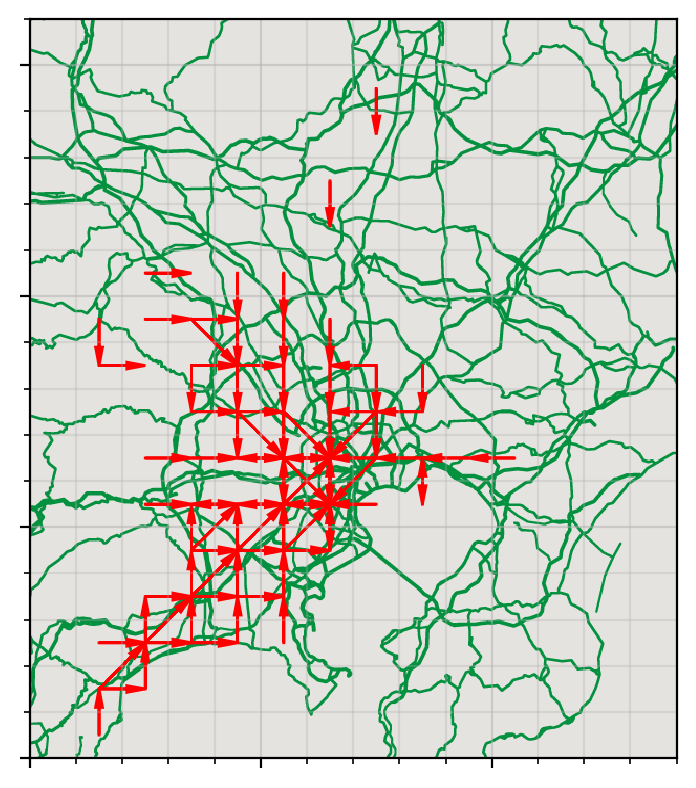}
    \end{subfigure}
    
    \begin{subfigure}{.2\textwidth}
    \centering
    \includegraphics[scale=0.27]{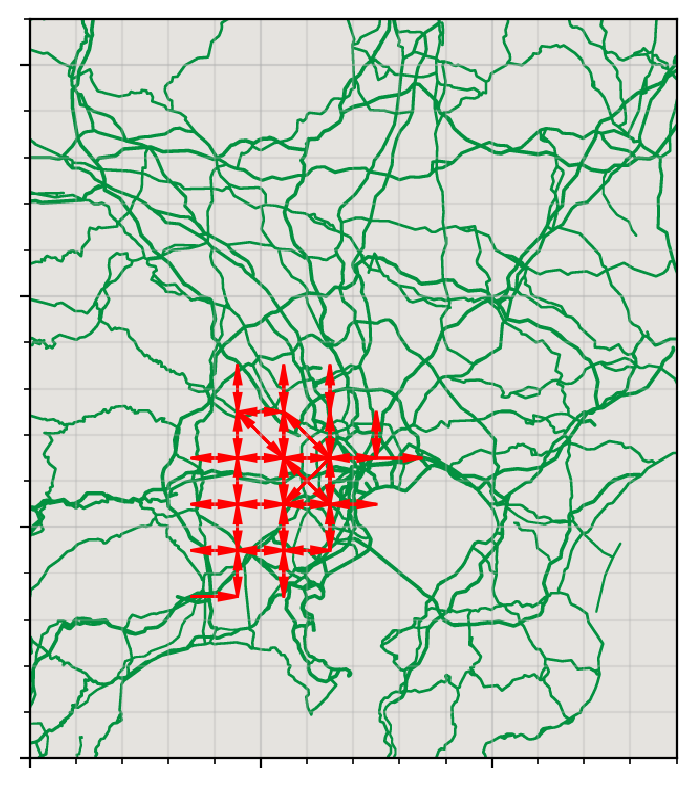}
    \end{subfigure}
    \begin{subfigure}{.2\textwidth}
    \centering
    \includegraphics[scale=0.27]{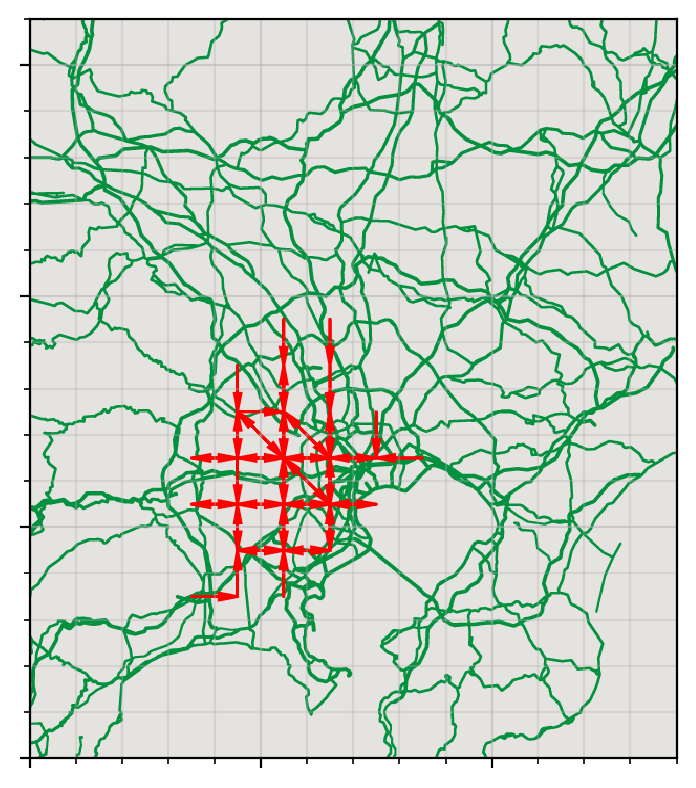}
    \end{subfigure}
    
    \begin{subfigure}{.2\textwidth}
    \centering
    \includegraphics[scale=0.27]{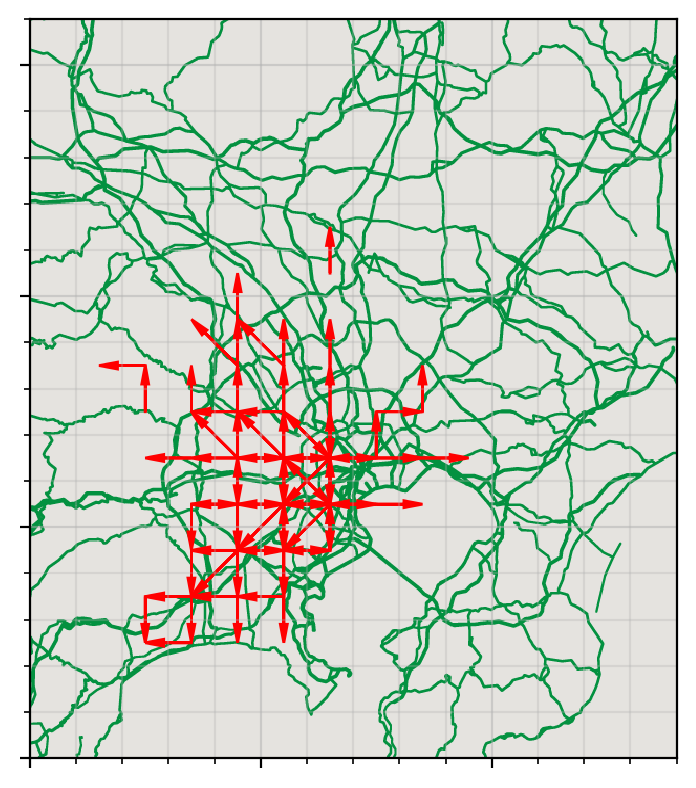}
    \caption{Estimation}
    \end{subfigure}
    \begin{subfigure}{.2\textwidth}
    \centering
    \includegraphics[scale=0.27]{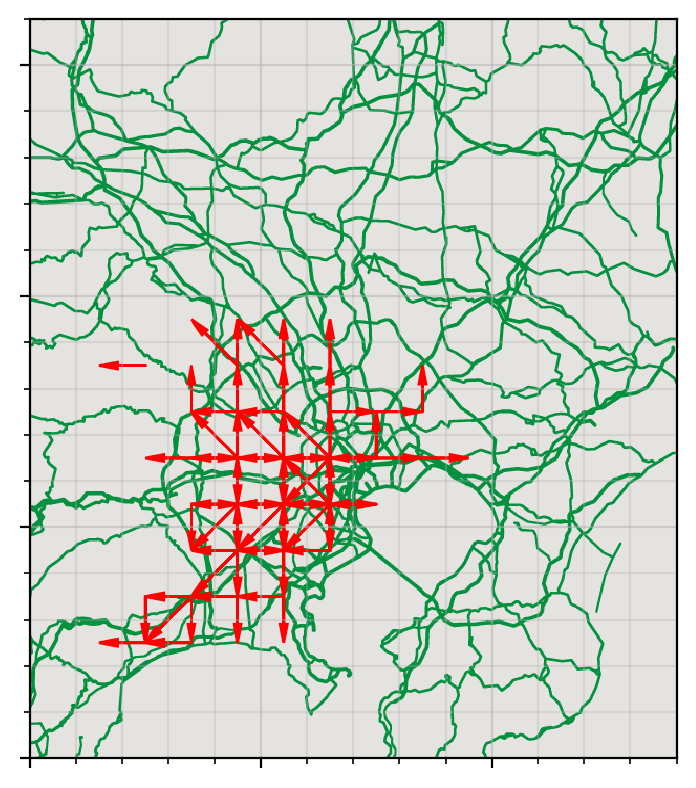}
    \caption{Ground Truth}
    \end{subfigure}
    \caption{Comparison between estimation based on our algorithm and ground truth movement. The four rows show the comparison at times 2:00, 8:00, 14:00 and 20:00, respectively. The red arrow depicts that flow between two block exceeds a threshold, 35.}
    \label{fig:city}
\end{figure}
    
    
\subsection{Estimating Spatio-Temporal Population Flow}
\label{subsec:flow}
We now turn to real-world aggregated data of population flow within the Japanese city of Tokyo. The dataset\footnote{Data Source: SNS-based People Flow Data, Nightley, Inc., Shibasaki \& Sekimoto Laboratory, the University of Tokyo, Micro Geo Data Forum, People Flow project, and Center for Spatial Information Science at the University of Tokyo,http://nightley.jp/archives/1954} consists of anonymous individual trajectories containing latitude and longitude of each person over time. The individual locations were recorded over time via geo-tagged tweets. We discretize the whole city area into $14\times16$ blocks with each block representing a $15 km \times 15 km$ area, resulting in (hidden) state space dimension of $d = 224$. The observations are collected by aggregating the individual trajectories every 30 minutes. A total of 6,432, 9,166, 6,822, 10,134, 6,646, 10,338 trajectories were collected respectively on July 1, July  7, October 7, October 13, December 16 and December 29 in the year 2013. We assume that the observations are corrupted by Gaussian Noise. Additionally, with a small chance, a point in the center block can be categorized to eight neighbouring blocks incorrectly, which account for sensor inaccuracy. In Figure~\ref{fig:city_heat}, we show the aggregate observation at timestamps 2:00 and 14:00 generated from the noisy observation model. The observations are based on all the individual trajectories recorded on the previously mentioned six number of days at corresponding times ($N=49538$).

Our task is to estimate the transition probabilities characterizing the population flow at different times 2:00, 8:00, 14:00, and 20:00. The estimations at each time point are based on one and half hour window such that the underlying HMM has a length of $T=3$. The observations are aggregated based on a time-homogeneous HMM with length $T=3$ and aggregate size $M=20$.  We estimate the HMM parameters directly from the aggregated data. We consider the estimated parameters with $M=1$ as the ground truth while assuming that the observation noise model is known. Figure \ref{fig:city} depicts the comparison between our estimation and ground truth movement at the four timestamps. The red arrows in the figure implicitly represent the underlying transition probabilities multiplied by the total population $N=49538$. One can observe that our algorithm successfully recovers the underlying movement of population with noisy aggregate observations.
\section{Conclusion}
\label{sec:conclusion}
In this paper, we proposed an algorithm for learning the parameters of a time-homogeneous HMM from aggregate data. Our algorithm is based on a modified version of the EM algorithm, wherein we utilized the Sinkhorn belief propagation algorithm to infer the unobservable states. In contrast to the existing state-of-the-art algorithms that explicitly consider the aggregate observation noise, our algorithm employs the aggregate observation noise within the graphical model and due to which it is consistent with the standard Baum-Welch algorithm when aggregate data consists of only a single individual. Moreover, our algorithm enjoys convergence guarantees. We further extended our algorithm to incorporate continuous observations and presented estimates for Gaussian observation model. In this work, we have assumed that the HMMs are time-homogeneous, which restricts the modeling capability of the data. We plan to explore learning of time-varying HMMs in our future research.

\bibliographystyle{plain}        
\bibliography{main}           

\appendix

\section{Proof of Proposition~\ref{prop:Maximization_HMM}}\label{appendix:proof_thm3}
\begin{proof}
    The M-step in Algorithm \ref{alg:mm_cgm} for aggregate HMMs solves
%
    \begin{subequations}\label{eq:M_step_approx}
    \begin{align}
        \min_{\theta} ~~& \cF(\bn, \theta) \label{eq:M_step_approx_a}
        \\
       \mbox{subject to} ~~& \sum_{x_1} \pi (x_1) = 1,  \label{eq:M_step_approx_b} 
       \\
       & \sum_{x_{t+1}} p(x_{t+1} \mid x_t) = 1,  \label{eq:M_step_approx_c}
       \\
       & \sum_{o_t} p(o_t \mid x_t) = 1 , \label{eq:M_step_approx_d}
    \end{align}
    \end{subequations}
where $\theta=\{\pi(x_1), p(x_{t+1}| x_t), p(o_t | x_t)\}$ and $\cF(\bn, \theta)$ as in \eqref{eq:bethe}.

Let the Lagrange multipliers corresponding to the constraints \eqref{eq:M_step_approx_b}, \eqref{eq:M_step_approx_c}, and \eqref{eq:M_step_approx_d} be $\lambda, \nu$, and $\mu$ respectively. Then, the Lagrangian is
    \begin{align*}
    &\cL(\theta, \lambda, \nu, \mu) =  \cF(\bn,\theta)  - \sum_{x_t} \nu_{x_t}( \sum_{x_{t+1}} p(x_{t+1} \mid x_t) - 1)\\
    &- \lambda ( \sum_{x_1} \pi (x_1) - 1) - \sum_{x_t} \mu_{x_t}( \sum_{o_t} p(o_t \mid x_t) - 1).
    \end{align*}
Setting the derivatives of the Lagrangian with respect to the variables to zero, we get
    \[
        \frac{\partial\cL}{\partial\pi(x_1)}
        = - \frac{n_1(x_1) }{\pi(x_1)} - \lambda = 0,
    \]
    \[
        \frac{\partial \cL}{\partial p(x_{t+1} \mid x_t)}
        = - \sum_{t=1}^{T-1}  \frac{n_{t,t+1}(x_t,x_{t+1})}{p(x_{t+1} \mid x_t)} - \nu_{x_t} = 0,
    \]
    \[
        \frac{\partial \cL}{\partial p(o_{t} \mid x_t)}
        = - \sum_{t=1}^{T}  \frac{n_{t,t}(x_t,o_{t})}{p(o_{t} \mid x_t)} - \mu_{x_t} = 0.
    \]
    Solving above equations, in view of the constraints \eqref{eq:M_step_approx_b}-\eqref{eq:M_step_approx_c}-\eqref{eq:M_step_approx_d}, we obtain
    \begin{subequations}\label{eq:parameters1}
    \begin{eqnarray}
        && \pi(x_1) = n_1(x_1),  \label{eq:parameters1_a} \\
       && p(x_{t+1} \mid x_t) = \frac{\sum_{t=1}^{T-1} n_{t,t+1}(x_t,x_{t+1})}{\sum_{t=1}^{T-1} n_{t}(x_t)}, \label{eq:parameters1_b}  \\
       && p(o_{t} \mid x_t) = \frac{\sum_{t=1}^{T} n_{t,t}(x_t,o_{t})}{\sum_{t=1}^{T} n_{t}(x_t)}. \label{eq:parameters1_c}  
    \end{eqnarray}
    \end{subequations}
\end{proof}


\section{Proof of Proposition~\ref{prop:connection_Baum_Welch}}
\label{appendix:proof_prop_connection_Baum_Welch}
\begin{proof}
Here we only present proof for the statement related to Algorithms~\ref{alg:mm_cgm}. It can be easily extended to the other part of the theorem.
In case $M=1$, the aggregate observation $\by$ corresponds to a sequence of observations $\hat{o}_1, \hat{o}_2,\ldots,\hat{o}_T$. In particular, the aggregate observations take the form 
\begin{equation}
    y_{t}(o_t) = \delta(o_t - \hat{o}_t),
\end{equation}
where $\delta(\cdot)$ denotes the Dirac function. Then the messages in collective forward-backward algorithm coincide with the messages in standard forward-backward algorithm~\cite{SinHaaZha20} and take the following form
\begin{subequations}\label{eq:standard_forward_backward}
\begin{eqnarray}
    &&\alpha_t(x_t) \propto \sum_{x_{t-1}} p(x_t|x_{t-1}) \alpha_{t-1} (x_{t-1}) p(\hat{o}_{t-1} | x_{t-1}), \label{eq:standard_forward_backward1}  \\
    &&\beta_t(x_t) \propto \sum_{x_{t+1}} p(x_{t+1}|x_{t}) \beta_{t+1} (x_{t+1}) p(\hat{o}_{t+1} | x_{t+1}), \label{eq:standard_forward_backwar2} \\
    &&\gamma_t(x_t) = p(\hat{o}_t | x_{t}). \label{eq:standard_forward_backwar3} 
\end{eqnarray}
\end{subequations}
Using above messages, the required marginals can be estimated as
\begin{subequations}
    \begin{eqnarray}
        && n_t(x_t) \propto p(\hat{o}_t | x_{t}) \alpha_t(x_t) \beta_t(x_t), \\
       && n_{t,t+1}(x_t,x_{t+1}) \propto   \alpha_t(x_t) p(x_{t+1} | x_t) \beta_t(x_{t+1}) \nonumber \\
       && \qquad \qquad \qquad \qquad \qquad p(\hat{o}_t | x_{t})  p(\hat{o}_{t+1} | x_{t+1}), \\
	&&n_{t,t}(x_t, \hat{o}_t) = n_t(x_t).
    \end{eqnarray}
\end{subequations}
Finally, the parameter update equations given in Algorithm~\ref{alg:mm_cgm} reduce to the standard Baum-Welch algorithm.
\end{proof}

\section{Proof of Proposition~\ref{prop:Maximization_HMM_cts}}
\label{appendix:proof_prop_cts}
\begin{proof}
In case of continuous (Gaussian) emission densities, the observations constitute $\bo = \{ o_1^{(m)}, \ldots , o_T^{(m)}\}$, $m=1,2,\ldots,M$ with $o_t^{(m)} $ being the continuous observation of $m$-th individual at time $t$. 

The M-step of the learning problem solves
    \begin{subequations}\label{eq:M_step_cts_approx}
    \begin{align}
        \min_{\theta }~~ & \cF(\bn,\theta) \label{eq:M_step_cts_approx_a}
        \\
      \mbox{subject to}~~ & \sum_{x_1} \pi (x_1) = 1,  \label{eq:M_step_cts_approx_b} \\
      &\sum_{x_{t+1}} p(x_{t+1} \mid x_t) = 1. \label{eq:M_step_cts_approx_c}
    \end{align}
    \end{subequations}
where $ \theta = \{\pi(x_1), p(x_{t+1} | x_t), \mu(x_t), \Sigma(x_t) \}$. The free energy $\cF$ is basically the same as in \eqref{eq:bethe}; the only difference is that $p(o_t|x_t)$ is a Gaussian parametrized by \eqref{eq:Gaussian}.

Let the Lagrange multipliers corresponding to the constraints \eqref{eq:M_step_cts_approx_b} and \eqref{eq:M_step_cts_approx_c} be $\lambda$ and $\nu$ respectively, then the Lagrangian is
    \begin{align*}
    & \cL(\theta, \lambda, \nu) =  \cF(\bn,\theta) - \lambda ( \sum_{x_1} \pi (x_1) - 1) \\
          &\quad \quad \quad \quad - \sum_{x_t} \nu_{x_t} ( \sum_{x_{t+1}} p(x_{t+1} \mid x_t) - 1).
    \end{align*}
Setting the derivatives of the Lagrangian with respect to $\{\pi(x_1), p(x_{t+1} | x_t)\}$ to $0$ we obtain
    \[
        \frac{\partial\cL}{\partial \pi(x_1)}
        =  \frac{n_1(x_1)}{\pi(x_1)} - \lambda = 0,
    \]
    \[
        \frac{\partial \cL}{\partial p(x_{t+1} \mid x_t)}
        = \sum_{t=1}^{T-1}  \frac{n_{t,t+1}(x_t,x_{t+1})}{p(x_{t+1} \mid x_t)} - \nu_{x_t} = 0.
    \]
Solving above equations, we arrive at
    \begin{subequations}\label{eq:parameters1_cts}
    \begin{eqnarray}
        && \pi(x_1) = n_1(x_1),  \label{eq:parameters1_cts_a} \\
      && p(x_{t+1} \mid x_t) = \frac{\sum_{t=1}^{T-1} n_{t,t+1}(x_t,x_{t+1})}{\sum_{t=1}^{T-1} n_{t}(x_t)}. \label{eq:parameters1_cts_b}
    \end{eqnarray}
    \end{subequations}

 
To find the Gaussian density parameter $\mu(x_t)$, differentiating the objective $\cF(\bn,\theta)$ with respect to $\mu(x_t)$ and equating it to zero, we get
\[
  \sum_{t=1}^{T} \sum_{m=1}^{M} n_t^{(m)} (x_t) (o_t^{(m)} - \mu(x_t))  = 0,
\]
and therefore
\[
    \mu(x_t) = \frac{\sum_{t=1}^{T} \sum_{m=1}^{M} n_t^{(m)} (x_t) ~o_t^{(m)} } {\sum_{t=1}^{T} n_t (x_t)} .
\]
Similarly, the update \eqref{eq:parameters_cts_d} can be obtained by setting $\partial \cF/\partial \Sigma(x_t)$ to zero.
\end{proof}

\end{document}